\documentclass[conference]{IEEEtran}
\IEEEoverridecommandlockouts

\usepackage{cite}
\usepackage{amsmath,amssymb,amsfonts}
\usepackage{algorithmic}
\usepackage{graphicx}
\usepackage{textcomp}
\usepackage{xcolor}
\usepackage{hyperref}
\usepackage{tabularx}
\usepackage{multicol}
\usepackage{adjustbox}
\usepackage{pgfplots}
\def\BibTeX{{\rm B\kern-.05em{\sc i\kern-.025em b}\kern-.08em
    T\kern-.1667em\lower.7ex\hbox{E}\kern-.125emX}}
\begin{document}

\title{ForDigitStress: A multi-modal stress dataset employing a digital job interview scenario\\

\thanks{
}
}

\makeatletter
\newcommand{\linebreakand}{%
  \end{@IEEEauthorhalign}
  \hfill\mbox{}\par
  \mbox{}\hfill\begin{@IEEEauthorhalign}
}
\makeatother

\author{\IEEEauthorblockN{Alexander Heimerl}
\IEEEauthorblockA{\textit{Lab for Human-Centered AI} \\
\textit{Augsburg University}\\
Augsburg, Germany \\
alexander.heimerl@uni-a.de}
\and
\IEEEauthorblockN{Pooja Prajod}
\IEEEauthorblockA{\textit{Lab for Human-Centered AI} \\
\textit{Augsburg University}\\
Augsburg, Germany \\
pooja.prajod@uni-a.de}
\and
\IEEEauthorblockN{Silvan Mertes}
\IEEEauthorblockA{\textit{Lab for Human-Centered AI} \\
\textit{Augsburg University}\\
Augsburg, Germany \\
silvan.mertes@uni-a.de}
\and
\IEEEauthorblockN{Tobias Baur}
\IEEEauthorblockA{\textit{Lab for Human-Centered AI} \\
\textit{Augsburg University}\\
Augsburg, Germany \\
tobias.baur@uni-a.de}
\and
\IEEEauthorblockN{Matthias Kraus}
\IEEEauthorblockA{\textit{Lab for Human-Centered AI} \\
\textit{Augsburg University}\\
Augsburg, Germany \\
matthias.kraus@uni-a.de}
\and
\IEEEauthorblockN{Ailin Liu}
\IEEEauthorblockA{\textit{Lab for Human-Centered AI} \\
\textit{Augsburg University}\\
Augsburg, Germany \\
ailin.liu@rwth-aachen.de}
\and
\IEEEauthorblockN{Helen Risack}
\IEEEauthorblockA{\textit{Lab for Human-Centered AI} \\
\textit{Augsburg University}\\
Augsburg, Germany \\
helenrisack@yahoo.de}
\and
\IEEEauthorblockN{Nicolas Rohleder}
\IEEEauthorblockA{\textit{Department of Psychology} \\
\textit{Friedrich-Alexander University}\\
Erlangen-Nürnberg, Germany \\
nicolas.rohleder@fau.de}
\linebreakand
\IEEEauthorblockN{Elisabeth Andr\'{e}}
\IEEEauthorblockA{\textit{Lab for Human-Centered AI} \\
\textit{Augsburg University}\\
Augsburg, Germany \\
elisabeth.andre@uni-a.de}
\and
\IEEEauthorblockN{Linda Becker}
\IEEEauthorblockA{\textit{Department of Psychology} \\
\textit{Friedrich-Alexander University}\\
Erlangen-Nürnberg, Germany \\
linda.becker@fau.de}
}

\maketitle

\begin{abstract}
We present a multi-modal stress dataset that uses digital job interviews to induce stress. The dataset provides multi-modal data of 40 participants including audio, video (motion capturing, facial recognition, eye tracking) as well as physiological information (photoplethysmography, electrodermal activity). In addition to that, the dataset contains time-continuous annotations for stress and occurred emotions (e.g. shame, anger, anxiety, surprise). In order to establish a baseline, five different machine learning classifiers (Support Vector Machine, K-Nearest Neighbors, Random Forest, Long-Short-Term Memory Network) have been trained and evaluated on the proposed dataset for a binary stress classification task. The best-performing classifier achieved an accuracy of 88.3\% and an F1-score of 87.5\%.
\end{abstract}


\section{Introduction}
Stress is the body's response to any demand or threat \cite{APA1}. It is a normal physiological reaction to perceived danger or challenge, and it can be beneficial in small doses, e.g it can support improving your performance or memory functions \cite{yaribeygi2017impact}\cite{becker2019time}. However, chronic stress can have a negative impact on both physical and mental health. Chronic stress can lead to a variety of mental health problems, including anxiety and depression. It can also make existing mental health conditions worse. Stress can lead to changes in brain chemistry and function, which can disrupt normal communication between brain cells. This can lead to symptoms such as difficulty concentrating, memory problems, and irritability. Stress can also lead to physical symptoms such as headaches, muscle tension, and fatigue \cite{APA2}. Stress can also impact physical health by weakening the immune system, increasing the risk of heart disease, and promoting unhealthy behaviors such as overeating, smoking, and drinking alcohol \cite{APA2}.

Among the many sources of stress, work-related stress is one of the most widespread and often seems inevitable. Therefore, there is a need to understand stressful situations at work and provide coping mechanisms on how to deal with such in order to prevent chronic stress. Especially, job interviews have been identified as one of the major stressors in a work-related context for many reasons. They often involve a lot of uncertainty, pressure, and potential rejection. In research, job interview scenarios have recently become a popular use case for studying how to recognize and regulate stress as a result of being a natural stress inducing event \cite{heimerl2022pupillometry, baur:2013:jobinterviews,heimerl2022generating}.

As remote job interviews have become common practice in response to the restrictions created by the SARS-CoV-2 crisis, such a setting has been used for collecting a novel multi-modal stress data set in a realistic naturally stress-inducing environment. For data collection, we recorded signals from various sources including audio, video (motion capturing, facial recognition, eye tracking) as well as physiological information (photoplethysmography (PPG), electrodermal activity (EDA)). Synchronization of the different signals was provided by the Social Signal Interpretation (SSI) framework \cite{wagner2013SSI}. We gathered data from 40 participants who took part in remote interview sessions, resulting in approximately 56 hours of multi-modal data. For data annotation, participants self-reported stressful situations during the interview as well as their perceived emotions. In addition, two experienced psychologists annotated the interviews frame-by-frame using equal stress and emotion labels. Calculating the inter-rater reliability for the individual labels resulted in substantial to almost perfect agreement (Cohen's $\kappa > 0.7$ for all labels). In addition to that, salivary cortisol levels were assessed in order to investigate whether the participants experienced a biological stress response during the interviews. 

For automatically classifying the participant's stress level during the interview, the collected signal information was used to produce a rich high-level feature set. The set contains EDA, HRV, body keypoints, facial landmarks including action units, acoustic frequency, and spectral features. Further, a pupil feature set was created based on the latent space features of an autoencoder that has been trained on close-up videos of the eye. In addition to that, the pupil diameter has been extracted as well. 
For reducing the dimensionality of the input feature vector, two approaches - early and late PCA (Principal Component Analysis) - were compared. The classification problem was formulated as a binary stress recognition task ((stress vs. no stress). We used and compared the performance of five different machine-learning classifiers - SVM (Support Vector Machine), KNN (K-Nearest Neighbors), NN (Feed-forward Neural Network), RFC (Random Forest Classifier), and LSTM (Long-Short-Term Memory Network) solely using the pupil features as input. To the best of our knowledge, we present the first approach utilising close-up eye features for detecting stress. Evaluation of the classifiers revealed that a NN approach using all modalities as input and applying early PCA led to the best recognition of the participant's stress level. An NN approach also performed best for each individual modality. Comparing the different modalities for their impact on the recognition performance, HRV features had the highest accuracy and $F_{1}$-scores. 

The proposed dataset makes the following contributions to the research community.
First, we provide data collected in a realistic stress setting that has been validated by the analysis of saliva cortisol levels in order to assess whether participants experienced a biological stress response during the interviews.
Secondly, the data set was annotated using a continuous labelling approach enabling dynamic stress recognition. Thirdly, we provide a multi-modal stress dataset containing established as well as novel modalities, e.g. close-up eye features

for providing a promising non-invasive modality for detecting stress. 
The structure of this article is as follows: In Section 2, we present background and related work regarding existing stress data sets.

Section 3 describes the data collection process including design principles, the recording system, properties of the data set as well as annotation procedure and feature extraction methods. The method for automatic stress recognition is explained in detail in Section 3. The results of the performance of the different machine-learning classifiers are presented in section 4 and discussed in Section 5. Finally, conclusions are provided in Section 6, and ethical considerations are detailed in Section 7.
\section{Background and related work}

As of today, multiple stress datasets for the automatic recognition of stress are available. 
\autoref{tab:stress_datasets} displays an overview of some of the existing stress datasets. 
The datasets not only differ in the used modalities for stress recognition but also in the stimulus to induce stress. 
Those stimuli range from realistic real-world scenarios to highly optimized lab settings. 

A common way to induce stress in a controlled way is to use established stress tests like the \emph{Trier Social Stress Test} (TSST).
The WESAD corpus by Schmidt et al. \cite{schmidt2018wesad} uses the TSST as a stimulus and provides physiological data. 
Further, various stress-related annotations like the affective state (neutral, stress, amusement) are given that were obtained by a variety of self-report questionnaires, e.g., Positive Negative Affect Schedule (PANAS; \cite{krohne1996untersuchungen}, State-Trait Anxiety Inventory (STADI; \cite{laux2013state}), SAM, Short Stress State Questionnaire and others.

Similarly, the UBFC-Phys dataset introduced by Sabour et al. \cite{sabour2021ubfc} used an approach inspired by the TSST to induce stress.
While also providing physiological data, that dataset contains stress states derived from pulse rate variability and EDA.

For the \emph{Multimodel Dataset for Psychological Stress Detection} (MDPSD) corpus provided Chen et al. \cite{chen2020introducing}, stress was induced by the classic Stroop Color-Word Test, the Rotation Letter Test, the Stroop Number-Size Test and the Kraepelin Test. 
Facial videos, PPG and EDA data are provided.
Stress annotations were obtained through a self-assessment questionnaire.

Koldjik et al. \cite{koldijk2014swell} introduced the SWELL dataset where they tried to simulate stress-inducing office work by applying time pressure in combination with typical work interruptions like emails.
In order to assess the subjective experience during the study they relied on various validated questionnaires to gather data about task load, mental effort, emotion response and perceived stress.

In contrast to the above datasets, Healey and Picard \cite{healey2005automobilestress} presented a dataset for \emph{Stress Recognition in Automobile Drivers} using highly realistic real-world stressors instead of rather controlled approaches to induce stress.
Here, they induced stress by letting the subjects perform open-road drivings. 
Besides recording physiological data, stress annotations were obtained through self-assessment questionnaires using free scale and forced scale stress ratings.

The MuSE dataset introduced by Jaiswal et al. \cite{jaiswal2020muse} also used a real-world stressor, but in contrast to other datasets, they did not induce stress by simulating a specific scenario themselves. 
They made use of the final exams period at a university as an external stressor.
Therefore, they recruited 28 college students and recorded them in two sessions, one during the finals period and one afterwards. 
During the recordings, they confronted the participants with various emotional stimuli. 
Afterwards, the participants self-reported their perceived stress and emotions.
Moreover, additional emotion annotations have been created by employing Amazon Mechanical Turk workers. 

Similar to the MuSE dataset the SWEET study \cite{smets2018sweetstudy} also relied on naturally occurring external stressors as a stimulus. 
They assessed stress manifested by daily life stressors of 1002 office workers. 
In contrast to other studies that are conducted in laboratory settings, they investigated the perceived stress of the participants during their daily life for five consecutive days. 
Throughout those five days, they collected physiological data with wearables, contextual information (e.g., location, incoming messages) provided by a smartphone as well as self-reported stress. 
The daily self-assessment was done with a smartphone application that questions the user 12 times a day about their perceived stress.

Further datasets exist that are based on non-physiological feature sets. 
For example, the Dreaddit corpus presented by Turcan et al. \cite{turcan2019dreaddit} contains a collection of social media posts that were annotated regarding stress by Amazon MTurk workers.

Other datasets, like the SLADE dataset introduced by Ganchev et al. \cite{ganchev2017overall}, focus on scenarios facilitating stress, although not explicitly giving stress annotations.
Instead, the SLADE dataset provides valence-arousal labels for situations where stress was induced using audio-visual stimuli, i.e. movie excerpts.

Similarly, the CLAS corpus presented by Markova et al. \cite{markova2019clas} provides valence-arousal labels as well as cognitive load annotations to situations where stress was induced by a math problems test, a Stroop test, and a logic problems test. 
Additionally, physiological data, such as ECG, PPG and EDA is provided.

Altogether, a large variety of different stress datasets already exists and is available to the research community. However, existing datasets show some drawbacks regarding stress labels, recorded modalities and availability. 

Existing stress datasets predominantly are labelled through stress questionnaires or similar assessments. Those approaches come with the disadvantage of yielding annotations of low temporal resolution, i.e., large time frames are treated as one and aggregated to a single annotation. 
As such, short-term deviations in stress levels can not be modelled with sufficient precision.
In contrast to that, the dataset presented in this paper was annotated in a time-continuous manner. This allows for the development of stress recognition systems that are more accurate, reactive and robust than is the case with existing datasets.

Even though multi-modal stress datasets exist they rarely provide a comprehensive representation of the participants' behavior. The majority of the datasets rely on physiological signals e.g. HRV and EDA with some of them also providing video or audio or already extracted features. However, to the best of our knowledge, there is no dataset present that provides a comprehensive collection of relevant modalities. The proposed ForDigitStress dataset contains audio, video, skeleton data, facial landmarks including action units as well as physiological information (PPG, EDA). In addition to the raw signals, we also provide already extracted features for HRV and EDA as well as established feature sets like GEMAPS\cite{gemaps} and OpenPose\cite{openpose}. Furthermore, this dataset contains pupillometry data which is a mostly overlooked modality for the recognition of stress. As prior work suggests \cite{partala2003pupil, sirois2014pupillometry, bradley2008pupil}, there are correlations between various affective states and pupil dilation. Also collecting pupillometry data can be done unobtrusively by using existing eyetrackers or even laptop webcams \cite{heimerl2022pupillometry}. Therefore, we believe that incorporating pupillometry data can benefit multiple stress-related use cases where eye-tracking is a reasonable option. The dataset provides already extracted pupil diameter as well as close-up infrared videos of the eye. Based on the close-up videos we trained an autoencoder and extracted the latent space features that represent an abstracted version of the eye. Those features are also made available as part of the dataset.

\section{Dataset}

\begin{table*}[]
\begin{tabularx}{\linewidth}{ XXXXXX }
\hline
Name & Number of participants & Duration per participant & Stress stimulus & Modality & Annotation \\ \hline
SWELL \cite{koldijk2014swell} & 25 & 3h & Office work with time pressure and email interruptions & HR, HRV, SC, Facial Expressions (Head Orientation, Facial Movements, Action Units, Emotion), Body Posture, Computer Interaction & Task NASA-TLX, RSME, Self-Assessment Manikins (SAM), Perceived Stress (10 point scale)\\ \hline
WESAD \cite{schmidt2018wesad} & 15 & 2h & TSST & Blood Volume Pulse, Electrocardiogram, Electrodermal Activity, Electromyogram, Respiration, Body Temperature, Three-axis Acceleration & Three different affective states (neutral, stress, amusement), PANAS, State-Trait Anxiety Inventory(STAI), SAM, Short Stress State Questionnaire(SSSQ), Additional assessment of Stressed, Frustrated, Happy and Sad  \\ \hline
MuSE \cite{jaiswal2020muse} & 28 & 45min & Final exams period & HR, SC, Breathing Rate, Skin Temperature, Audio,  Video, Thermal Recordings of the Face & Perceived Stress
Scale, SAM, Big-5 personality scores  \\ \hline
CLAS \cite{markova2019clas} & 62 & 30min & Math problems test, Stroop test, Logic problems test & ECG, PPG, EDA, Three-axis Acceleration & Cognitive load, Valence-Arousal,  \\ \hline
UBFC-Phys \cite{sabour2021ubfc} & 56 & 9min & Based on TSST (speech task, arithmetic task) & BVP, EDA, Video & CSAI (cognitive anxiety, somatic anxiety, self-confidence) \\ \hline
Stress Recognition in Automobile Drivers \cite{healey2005automobilestress} & 24 & $>$50min & Open road driving & ECG, EMG, EDA, Breathing Rate & Free scale stress rating (1-5), forced scale stress rating (1-7)
\\ \hline
SWEET study \cite{smets2018sweetstudy} & 1002 & continuous monitoring for 5 consecutive days & Daily life & ECG, SC, Skin Temperature, Three-axis Acceleration & PSS, PSQI, DASS, RAND-36 self-reported stress through ecological momentary assessments (EMAs), Leuven Postprandial Distress Scale, SAM
\\ \hline

MDPSD \cite{chen2020introducing} & 120 & $<$4min & Stroop Color-Word Test, Rotation Letter Test, Stroop Number-Size Test, Kraepelin Test & Facial videos, PPG, EDA & Self-assessment questionnaire
\\ \hline

\end{tabularx}
\vspace{0.1cm}
\caption{Overview of
existing stress datasets}
\label{tab:stress_datasets}
\end{table*}

\subsection{Design Principles}
\subsubsection{Setting}
The main requirement for the setup has been to elicit stress and emotional arousal in participants. Moreover, the setting should reflect a familiar real-world scenario. Therefore, we opted for a remote job interview scenario, a typical digital stressor. Performing remote job interviews has become a common procedure in many modern working environments. Job interviews are by their nature a complex stressful social scenario where different aspects of human interaction and perception collude. Previous research has shown that psycho-social stress also occurs in mock job interviews \cite{campisi2012:mock_interview_stress, gebhard2014exploring_mock_interview}. \autoref{fig:study_setup} shows a schematic of the employed study setup. To mimic remote job interviews participant and interviewer were interacting via two laptops while sitting in two separate rooms.
\subsubsection{Procedure}
Participants were invited to the laboratory and were told that physiological reactions during an online job interview will be recorded. 
In advance, participants sent their curriculum vitae (CV) to the experimenter and filled out an online survey, in which demographic variables and experiences with job interviews were assessed. 
After arrival in the laboratory, they were asked about their dream job and were equipped with PPG and EDA sensors as well as a wearable eye tracker. 
Then, they had fifteen minutes to prepare for the interview. The participant and interviewer were seated in two separate rooms and were interacting with each other over two connected laptops similar to an online meeting.
The interviewer tried to ask critical questions to stress the applicant and to induce negative emotions. 
Contents of the interviews included questions about strengths and weaknesses of the applicant, dealing with difficult situations on the job, salary expectations, willingness to work overtime, as well as inconsistencies in the CV. In addition, tasks related to logical thinking were asked as well as questions about basic knowledge in the areas of mathematics and language. The procedure is described in detail in \cite{becker2023_fordigitstress}. After the job interviews, participants were asked about their emotions during the interview. 
After this, participants reported whether they felt stressed at any time during the interviews. 
Afterwards, participants were instructed to describe as precisely as possible in which specific situations during the job interviews they felt stressed. 
This procedure (rating and assignment to specific situations) was repeated for all of the reported emotional states (i.e., shame, anxiety, pride, anger, annoyed, confused, creative, happy, insecure, nervous, offended, sad, surprised). \\
In order to assess whether the mock job interview did elicit stress in the participants not only self-reports were collected but also saliva samples were taken to determine cortisol levels. Salivary cortisol levels are a measure for the activity of the hypothalamic-pituitary adrenal (HPA) axis. Increased cortisol levels can be observed when a person is exposed to stress \cite{bozovic2013salivarycortisol}, especially for social-evaluative situations. They are therefore an adequate measure to investigate the participant's biological response to the remote job interview. When a person has been exposed to a stressor the cortisol level does not increase instantly. Peak levels are usually found about 20 minutes after psycho-social stressors of short duration (e.g., the TSST). After this, cortisol levels return to baseline levels. The samples of participants that have been stressed by the interview will show an increase in cortisol level until they reach a peak followed by a decrease back to their baseline levels. Therefore, salivary cortisol was assessed as a measure for biological stress. For saliva collection, Salivettes (Sarstedt, Numbrecht) were used. Each participant provided six saliva samples at different time points. 
\autoref{fig:saliva_samples} displays an overview of the timing of saliva sample collection during the study. The first sample was collected at the beginning of the study and the second at the end of the preparation phase (i.e., immediately before the actual job interview started). Those two samples were separated by about 15 minutes in order to assess the baseline cortisol level before the participant was exposed to the stressor, i.e., the job interview. The further four samples were collected immediately after the job interview, 5 minutes, 20 minutes, and 35 minutes after it to cover the cortisol increase, its peak, and its return to baseline. 
During each saliva sampling, participants rated their current stress level on a 10-point Likert scale with the anchors "not stressed at all" and "totally stressed". 

\begin{figure}[ht]
\centering
\includegraphics[width=0.49\textwidth]{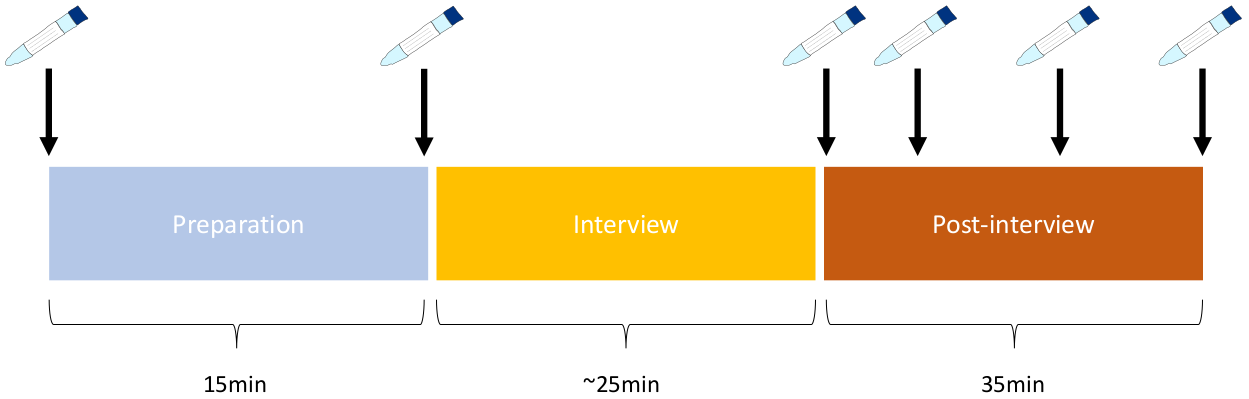}
\caption{Overview of the timing of saliva sample collection during the different stages of the study.}
\label{fig:saliva_samples}
\end{figure}

\subsection{Recording System}
Various sensors were used to record the participants' physiological responses. For recording and streaming the participant's data, we employed a Microsoft Kinect 2. The Microsoft Kinect 2 supports FullHD video captures as well as optical motion capturing to extract skeleton and facial data. Moreover, the built-in microphone was used to record ambient sound data. In addition to that, the participants were equipped with an ordinary USB headset from Trust. Furthermore, the IOM-biofeedback sensor was used to collect PPG and EDA data. Finally, participants were wearing a Pupil Labs eyetracker to record closeup videos of their eye. All sensors were connected to a Lenovo Thinkpad P15. The setup for the interviewers only consisted of audio recorded with the same Trust USB Headset and video from the built-in Lenovo Thinkpad P15 webcam. A schematic overview of the recording setup is displayed in \autoref{fig:study_setup}. The participant and interviewer were seated in two different rooms and were interacting remotely with each other through the two laptops. In a third room, another computer was set up to act as an observer. This way the interaction between the participant and the interviewer could be monitored unobtrusively. In order to keep the recorded signals in synchrony we implemented a SSI \cite{wagner2013SSI} pipeline. 
\begin{figure*}[ht]
\centering
\includegraphics[width=1\textwidth]{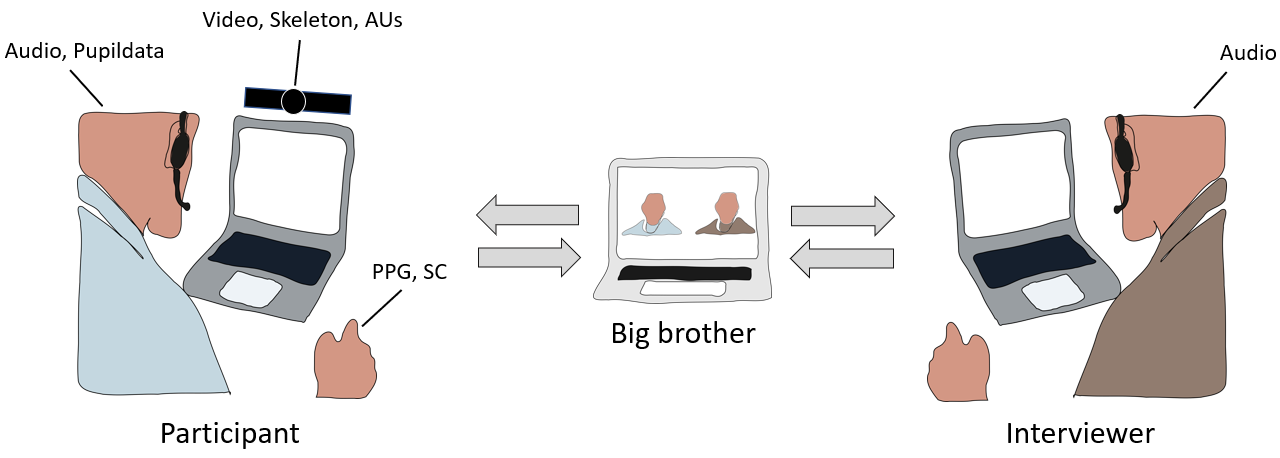}
\caption{Overview of the study setup. Participant and interviewer were seated in different rooms and interacting remotely with each other. A third computer was acting as an observer to unobtrusively monitor the interaction between the participant and interviewer.}
\label{fig:study_setup}
\end{figure*}

\subsection{Collected Data}
Data of \textit{N} = 40 healthy participants (57.5\% female, 40\% male, 2.5\%  diverse) was included in the data set. Mean age was $22.7 \pm 3.2$ years (min: 18, max: 31). Mean body-mass-index (BMI) was $23.2 \pm 4.1$  $kg/m^{2}$ (min: 17.9, max: 37.7; 1 missing). In total 56 hours and 24 minutes of multi-modal data have been recorded. An overview of all the recorded files is displayed in \autoref{files_table}.

\begin{table}[]
\centering
\begin{tabular}{lll}
\hline
\textbf{Sensor} &\textbf{Filename} & \textbf{Signal} \\ \hline
IOM & bvp.stream & PPG \\
& sc.stream & EDA \\ \hline
Kinect & video.mp4 & HD Video \\
 & skel.stream & Skeleton Data \\
 & face.stream & Facial Points \\
 & head.stream & Head Position \\
 & au.stream & Action Units \\
 & kinect.wav & Audio (room) \\ \hline
Headset & close.wav & Audio (close-talk)  \\ \hline
Eyetracker & eye.mp4 & Video (close up eye) \\ \hline \end{tabular}
\vspace{0.1cm}
\caption{\label{files_table} List of recorded files available for download}
\end{table}
 
\subsection{Annotation}
The basis for the annotations was the self-reports of the participants regarding perceived stressful situations and emotions. Two experienced psychologists annotated the recorded sessions frame by frame based on the participants' reports and content of the interviews. Categories for the annotations were the categories from the questionnaire (i.e., stress as well as the reported emotions like shame, anxiety, anger, and pride). In total, 1,286 minutes of data were annotated. There were no disagreements between the psychologists' ratings and the participants' self-reports, i.e., for every situation that was assigned to stress or an emotion by the participants, a time window could be assigned by the psychologists and a corresponding annotation could be created. \autoref{fig:label_distribution} displays the overall label distribution for the occurred emotions.

In the first step, the two psychologists independently annotated the 40 videos with the NOVA tool \cite{heimerl:2019:nova}. During the annotation process, the job interview videos were examined with regard to stress and different emotions. The annotations were created based on the observable behavior of the participants.
In a second step, the annotations were supplemented with information from the self-reports of the interviewees. For every visible or reported feeling of stress, a discrete label was created. 
Emotions were annotated accordingly. In a last step, disagreements in the annotations were discussed by the two psychologists. The annotations in which there was agreement, regarding the subject's perception of stress were adjusted. Situations that continued to be interpreted differently after the discussion remained unchanged in the annotation.

\begin{figure}[ht]
\centering
\includegraphics[width=0.49\textwidth]{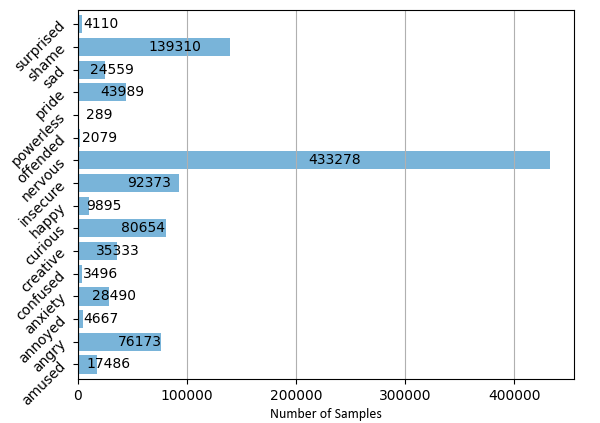}
\caption{Number of samples per occurred emotion.}
\label{fig:label_distribution}
\end{figure}

A screenshot of a loaded recording session from the dataset is shown in \autoref{fig:nova:overall}. In order to measure the quality and reliability of the discrete annotations we calculated the interrater agreement between the two psychologists using Cohen's Kappa (see \autoref{fig:interrateragreement}). The majority of the annotations have shown a strong to almost perfect agreement following the interpretation for Cohen's Kappa. 

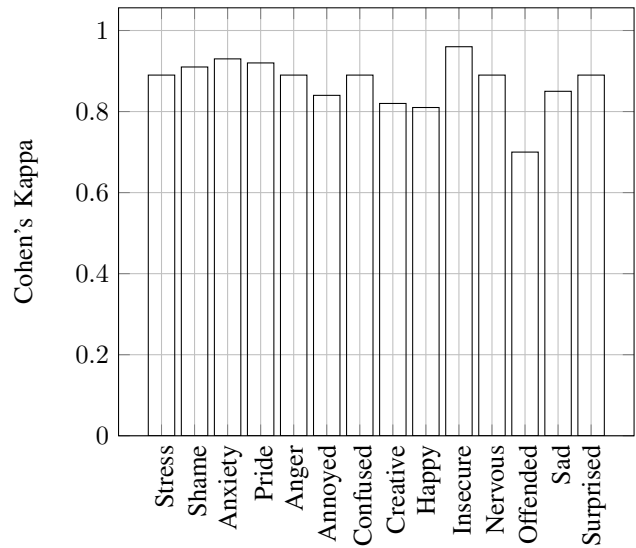
\begin{figure}
\centering
\begin{tikzpicture}
\begin{axis}[
symbolic x coords={Stress,Shame,Anxiety,Pride,Anger,Annoyed,Confused, Creative, Happy, Insecure,Nervous,Offended,Sad,Surprised},
xtick={Stress,Shame,Anxiety,Pride,Anger,Annoyed,Confused, Creative, Happy, Insecure,Nervous,Offended,Sad,Surprised},
xticklabel style={text height=2ex, rotate=90},
ymajorgrids=true,
xmajorgrids=true,
ymin=0,
ylabel=Cohen's Kappa] 
\addplot [ybar]
	coordinates {(Stress,0.89) (Shame,0.91)
		 (Anxiety, 0.93) (Pride, 0.92) (Anger, 0.89) (Annoyed, 0.84) (Confused, 0.89) (Creative, 0.82) (Happy, 0.81) (Insecure, 0.96) (Nervous, 0.89) (Offended, 0.70) (Sad, 0.85) (Surprised, 0.89)};
\end{axis}
\end{tikzpicture}
\caption{Average Cohen's Kappa calculated for stress and each emotion to map the interrater agreement between the two psychologists.}
\label{fig:interrateragreement}
\end{figure}

\begin{figure*}[ht]
\centering
\includegraphics[width=1\textwidth]{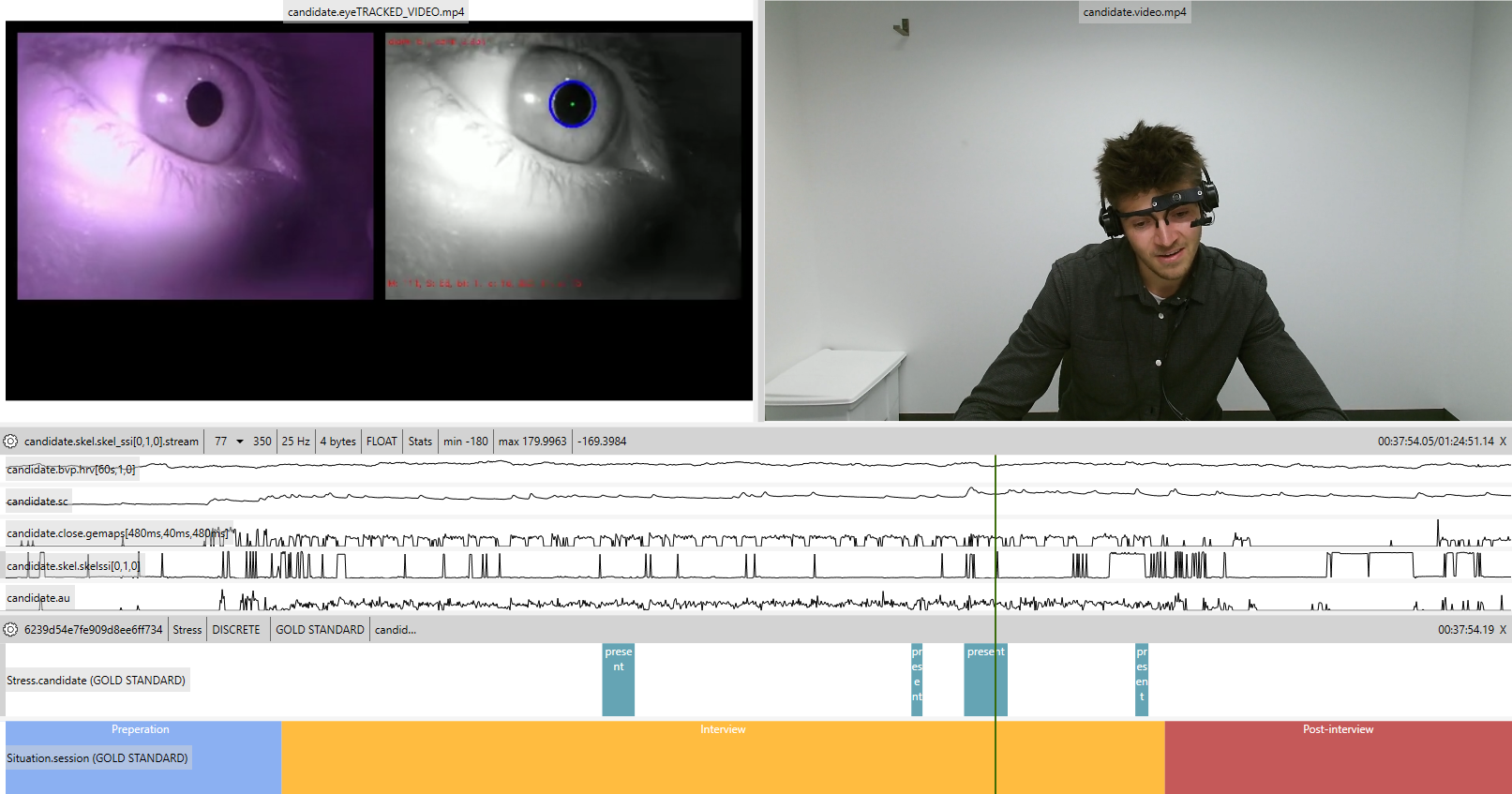}
\caption{An instance of a recorded session loaded in \textsc{NOVA}.The top row displays the eyetracking video alongside the video recording of the participant. Below that several feature streams are displayed: HRV feature stream, EDA, Gemaps audio features, skeleton data, and action units. At the bottom, two discrete annotation tiers are shown displaying stressful situations and the interview phase.}
\label{fig:nova:overall}
\vspace{-0.5cm}
\end{figure*}

\subsection{Feature Extraction}
The recorded raw data has been used to extract features that are valuable for stress recognition. The following section gives an overview of the extracted features as well as additional information regarding the extraction process. Moreover, the presented features are also available for download.

\subsubsection{EDA}
Features derived from the EDA signal are widely used for stress recognition~\cite{schmidt2018wesad, healey2005automobilestress, nkurikiyeyezu2019importance, koldijk2014swell}.
The EDA signal can be decomposed to skin conductance level (SCL) and skin conductance response (SCR)~\cite{schmidt2018wesad, setz2009discriminating}.
SCL or the tonic component is the slow-changing part of the EDA signal.
SCR or the phasic component is the rapid changes as a response to a specific stimulus.
First, we remove the high-frequency noise by applying a $5$ Hz low-pass filter~\cite{schmidt2018wesad, setz2009discriminating}.
We use the filtered signal to calculate statistical features~\cite{schmidt2018wesad, nkurikiyeyezu2019importance, sriramprakash2017stress} like mean, standard deviation, dynamic range, etc.
We compute the SCL and SCR components using the cvxEDA decomposition algorithm~\cite{greco2015cvxeda}.
In addition to the various statistical features of SCL and SCR signals, we also compute features derived from the peaks in the SCR signal~\cite{healey2005automobilestress}.
We compute a total of $17$ features (see~\autoref{tab:feats}) from a $60$ seconds long input EDA signal.

\subsubsection{PPG}
As demonstrated in previous studies~\cite{schmidt2018wesad, cho2018instant}, the PPG signal can be used to derive HRV (Heart Rate Variability) features for predicting stress.
We compute $22$ PPG-based HRV features which are listed in~\autoref{tab:feats}.
To derive the HRV from PPG, we detect the Systolic Peaks (P) from the input signal.
The first step is to remove baseline wander and high-frequency noises from the raw PPG signal.
We use a band-pass filter ($0.5-8$ Hz) to reduce the noise and enhance the peaks~\cite{elgendi2013systolic}.
Next, we use a peak finding algorithm to detect peaks such that (a) their amplitudes are above a specified threshold and, (b) consecutive peaks are sufficiently apart. 
The amplitude threshold is set to the mean of the $75$ percentile and $90$ percentile of the peak heights in the input signal.
The typical maximum heart rate of healthy participants during exercise stress is 3 beats per second (180 beats per minute)~\cite{kostis1982effect}.
Hence, we set the minimum time between two consecutive peaks as $0.333$ seconds.
We use $60$ seconds long PPG segments to detect the peaks and compute the HRV signal.
We compute various HRV features~\cite{schmidt2018wesad, pham2021heart, shaffer2017overview, sriramprakash2017stress, nkurikiyeyezu2019importance} from time domain, frequency domain, and Poincaré plots.

\subsubsection{Body keypoints}
Prior studies have established the value of body language and body behavior for the recognition of stress \cite{giakoumis2012behaviouralstressdetection}
\cite{aigrain2015bodylanguagestress}
\cite{chen2019gestureemotionalstress}. Therefore, our study setup included a Microsoft Kinect2 to extract 3D body data. This data provides information about 25 joints consisting of position in 3D space, orientation of the joints in 3D space as well as a confidence rating in regard to the tracking performance. Even though the Microsoft Kinect2 has been used in prior studies in the context of stress recognition \cite{aigrain2015bodylanguagestress, giakoumis2012behaviouralstressdetection, chen2019gestureemotionalstress} we aimed to provide additional body data in order to enable others to utilize the provided dataset across multiple datasets. Therefore, we extracted the OpenPose \cite{openpose} features from the recorded HD video displaying the participant. OpenPose is a widely used state-of-the-art framework for the detection of human body key points in single images. It is important to point out that OpenPose solely returns the body key points in 2D space, therefore, losing some information when compared to the Microsoft Kinect2 data. However, in order to extract the OpenPose features no special hardware is required and the data of a simple camera is sufficient enough. It is important to point out that due to the study setup, not all joints could be successfully tracked as the participants were sitting and their lower body was concealed by the table. Therefore only the features corresponding to the upper body joints provide reliable information.

\subsubsection{Action units}
Facial expressions play an important role in communicating emotions and therefore are frequently used for the automatic detection of affective states \cite{cohn2015automatedfaceanalysis}\cite{TARNOWSKI2017emotionrecognitionfacial}. Further, recent studies have utilized facial action units to successfully predict human stress\cite{aigrain2015bodylanguagestress}
\cite{giannakakis2020austress} \cite{giannakakis2022automaticstressau}. We extracted 17 facial action units (see \autoref{tab:feats}) provided by the Microsoft Kinect2. In addition to that, we also extracted the OpenFace2\cite{Baltruaitis2018OpenFace2F} features that consist of facial landmarks, head pose, facial action units, and eye-gaze information. Similar to OpenPose, those features can be extracted from any video data.

\subsubsection{Audio features}
Knapp et al.\cite{nonverbalcommunicationinhumaninteraction} argue that emotions are reliably transported by the voice. Indeed it is a well-established fact that the acoustic characteristics of speech e.g. pitch and speaking rate are altered by emotions\cite{tao2005affectivecomputingreview}. Moreover, vocal signs of stress are mainly induced by negative emotions \cite{lefter2016stressdetectionaudio}. Multiple studies were able to show that it is possible to automatically detect stress with acoustic features \cite{lu2012stressaudiosmartphone}\cite{lefter2016stressdetectionaudio}\cite{kurniawan2013stressspeech}\cite{han2018stressspeechdeeplearning}. In order to provide meaningful acoustic features we chose to extract the GEMAPS features \cite{gemaps}. One of the main objectives of the GEMAPS feature set has been to provide access to a comprehensive and standardized acoustic feature set. It contains frequency and energy-related features like pitch, jitter, shimmer and loudness, as well as spectral features, e.g., Hammarberg Index and harmonic differences. We calculated the features over a one second time window.

\begin{table*}[ht]
\caption{List of features extracted from various modalities}
\begin{center}
\begin{tabular}{|p{1.25cm}|p{5cm}|p{10cm}|}
\hline
\textbf{Modality}&\textbf{Features} & \textbf{Description} \\
\hline
Action Units & Jaw & Intensities of action units JawOpen, JawSlideRight\\
& Lips & Intensities of action units LipPucker, LipStretcherRight, LipStretcherLeft, LipCornerPullerLeft, LipCornerPullerRight, LipCornerDepressorLeft, LipCornerDepressorRight, LowerLipDepressorLeft, LowerLipDepressorRight\\
& Cheeks & Intensities of action units LeftCheekPuff, RightCheekPuff\\
& Eyes & Intensities of action units LeftEyeClosed, RightEyeClosed, RightEyebrowLowerer, LeftEyebrowLowerer\\
\hline
EDA & MeanEDA, StdEDA, MinEDA, MaxEDA, RangeEDA & Mean, Standard deviation, Min, Max, Dynamic Range of EDA signal \\
& SlopeEDA, MeanDeriv, StdDeriv & Slope of EDA signal, Mean and Standard deviation of 1st derivative of EDA signal \\
& MeanSCR, StdSCR, MeanSCL, StdSCL & Mean and Standard deviation of SCR and SCL components \\
& CorrSCL & Correlation of SCL with time \\
& PeaksSCR, AmplitudeSCR, DurationSCR, AreaSCR & Number of peaks, Sum of peak amplitudes, sum of peak durations and sum of area under the peaks of the SCR signal \\
\hline
PPG & HR & Number of peaks in $1$ minute\\
& MeanNN, MedianNN, MadNN & Mean, Median, Median absolute deviation of HRV \\
& StdNN, CVNN, IQRNN & Standard deviation, Coefficient of Variation, Inter-Quartile Range of HRV \\
& RMSSD, StdSD & Root Mean Square and Standard deviation of successive differences of P-P intervals \\
& pNN50, pNN20 & Percentage of successive differences of P-P intervals $> 50\ ms$ and $> 20\ ms$ \\
& TINN, HTI & Triangular Interpolation of HRV histogram, HRV Triangular Index \\
& LF, HF, LF/HF & Low Frequency ($0.04\ Hz - 0.15\ Hz$) and High Frequency ($0.15\ Hz - 0.4\ Hz$) power \\
& LFn, HFn & Normalized low and high-frequency power, LF/total power, HF/total power \\
& SD1, SD2, SD1/SD2 & Spread of HRV points on Poincaré plot along identity line and perpendicular to it \\
& S & Area of the ellipse formed by HRV points in the Poincaré plot \\
\hline

Audio & GEMAPS & Pitch, Jitter, Formant 1-3 frequency and relative energy, Formant 1 bandwidth, Shimmer, Loudness, Harmonics-to-noise ratio(HNR), Alpha Ratio, Hammarberg Index, Spectral Slope, Harmonic difference H1-H2, Harmonic difference H1-A3 \\ \hline
Body Keypoints & Kinect & x, y, z position and rotation of head, forehead, nose, left/right ear, chin, neck, torso waist, left/right shoulder, left/right elbow, left/right wrist, left/right hand, hip left/right, left/right knee, left/right ankle, left/right foot\\
& OpenPose & x, y position of nose, left/right eye, left/right ear, neck, left/right shoulder, left/right elbow, left/right wrist, hip left/middle/right, left/right knee, left/right ankle, left/right big toe, left/right small toe, left/right heel \\\hline 
Eye & Pupil features & pupil diameter, latent space features extracted from an autoencoder trained on the close-up videos of the eye \\  \hline

\end{tabular}
\label{tab:feats}
\end{center}
\end{table*}

\subsubsection{Pupil features}

Responses of the pupil like pupil dilation are closely related to subjective and physiological stress responses \cite{zekveld2019please} \cite{pedrotti2014automatic}.  
Furthermore, a recent study has shown that pupillometry is a suitable tool to measure arousal during emotion regulation after an acute stressor \cite{langer2021delayed} \cite{pedrotti2014automatic}. Therefore, part of our study setup has been a wearable eye tracker that provides close-up video data of the participant's eye. From those videos, we automatically extracted the pupil diameter by employing the extraction pipeline described in \cite{heimerl2022pupillometry}. In addition to that, we also trained an autoencoder on the close-up eye videos in order to extract the corresponding latent space features. The latent space features contain an abstract representation of the eye. \autoref{fig:autoencoder_transformation} displays the original input image of the eye and the output image produced by the autoencoder below. During the encoding and decoding process, barely any loss of information occurred as the input image and corresponding output image are almost identical. This is a strong indicator that the autoencoder has learnt meaningful features to accurately translate the image into and out of the latent space. The resulting feature set consists of 512 parameters corresponding to the size of the latent space. 

\begin{figure}[ht]
\centering
\includegraphics[width=0.49\textwidth]{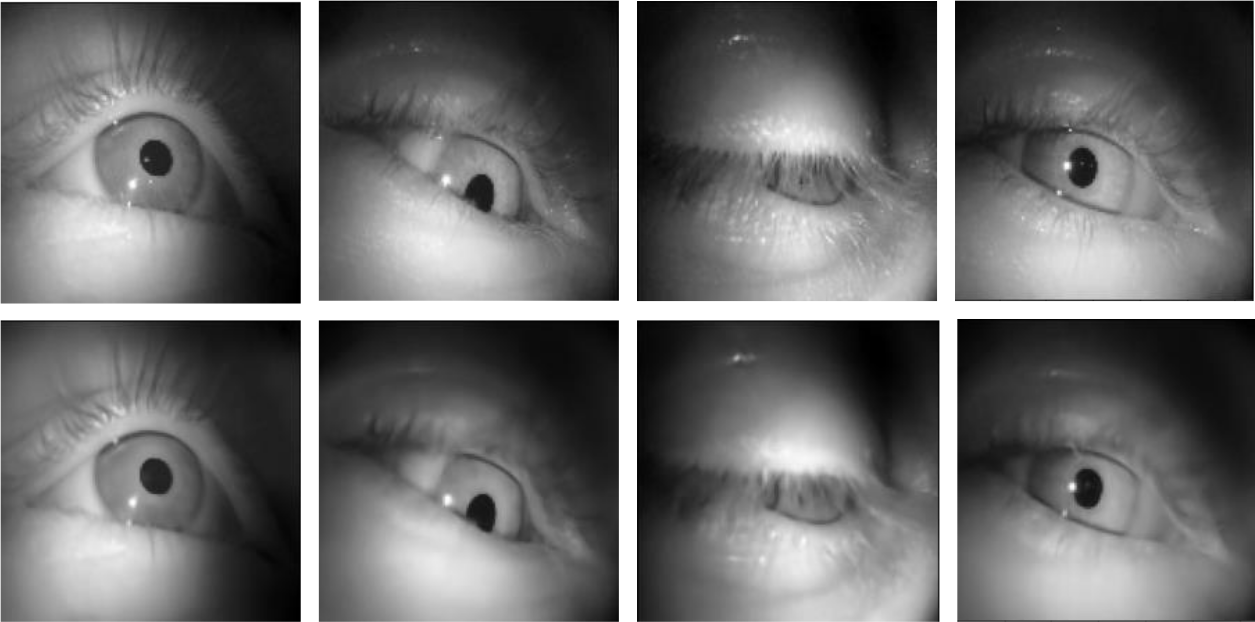}
\caption{Examples of reconstructed images. The top row displays the original input image, while the bottom row shows the images reconstructed with the autoencoder.}
\label{fig:autoencoder_transformation}
\end{figure}

\subsection{Availability}
The ForDigitStress dataset is freely available for research and non-commercial use. Access to the dataset can be requested at \url{https://hcai.eu/fordigitstress}. The dataset is organized in sessions with a total size of approximately 360 GB.

\subsection{Automatic Stress Detection}
\label{automatic_stress_detection}

\subsubsection{Dimensionality Reduction}
As seen from~\autoref{tab:feats}, various features have been extracted from each modality. 
The size of the input dimension can be a concern for some machine learning techniques, especially when we consider multi-modal stress recognition.
Therefore, we use PCA (Principal Component Analysis) as it has been shown to reduce dimensionality without a drop in classification performance of the machine learning models~\cite{reddy2020analysis}.
We apply PCA for stress models involving individual modalities as well as the multi-modal stress recognition models.
The length of the feature vectors of action units, EDA, HRV, OpenPose, and Gemaps was $17, 17, 22, 24, 58$.
We retain $95\%$ of the components using PCA, reducing the length of the feature vectors to $10, 9, 10, 8, 19$, respectively.
We follow two approaches for combining features for multi-modal stress recognition - early PCA and late PCA. 
In early PCA, we first apply PCA to individual modality features and then combine them.
Whereas in late PCA, we first combine the features and then apply PCA to the combined feature vector.
The length of the feature vector for early PCA is $56$ (sum of the length of feature vectors of each modality), and for late PCA is $49$.
Similar to Reddy et al.~\cite{reddy2020analysis}, we perform a MinMax normalization before applying PCA. 

\subsubsection{Classifiers}
Previous works~\cite{schmidt2018wesad, garg2021stress, sriramprakash2017stress, koldijk2016detecting} have demonstrated that many of the machine learning classifiers such as SVM (Support Vector Machine), KNN (K-Nearest Neighbors) and RFC (Random Forest Classifier) can achieve good stress recognition performance.
Recent works~\cite{bobade2020stress} have shown that simple feed-forward neural networks perform better than popular machine learning classifiers in feature-based stress recognition.
We train the following classifiers as a baseline for our dataset.

\begin{itemize}

\item \textbf{KNN}
This machine learning technique classifies samples based on the labels of the nearest neighbouring samples.
The neighbouring samples are determined using the Euclidean distance between them.
We use $K = 50$ neighbouring samples to classify the samples.

\item \textbf{Simple Neural Network}
This is a Multi-Layer Perceptron with an input layer, two hidden layers, and a prediction layer.
Since the size of the input varies depending on the modalities, we have a varying number of nodes in the hidden layers.
We set the number of nodes in the first hidden layer as half of the input size, rounded up to a multiple of $2$. 
The number of nodes in the second hidden layer is half of the first layer.
The activation function for hidden layers is ReLU (rectified linear unit).
The prediction layer has a single node with Sigmoid activation to discern between stress and no-stress classes.
We avoid over-fitting by using a dropout layer (dropout rate $=\ 0.2$) after the input layer. 

\item \textbf{RFC}
This is an example of an ensemble classifier that trains a number of decision tree classifiers on subsets of the training set.
This training technique controls over-fitting.
Hence, the RFC achieves better overall performance, even if the individual decision trees are weak.
In our evaluations, we use an RFC with $100$ decision trees (or estimators) and $50$ minimum samples for splitting a node.

\item \textbf{SVM}
This is a popular supervised learning technique that often achieves good stress recognition performance.
Similar to previous works~\cite{koldijk2016detecting, sriramprakash2017stress}, we use the Radial basis function (Rbf) as the kernel function for our SVMs.

\end{itemize}

The simple neural networks were implemented using Tensorflow. 
We use the SGD optimizer (learning rate $=\ 0.001$) and binary cross-entropy loss.
We train them for $100$ epochs with a batch size of $256$.
All other machine learning models were trained using Scikit-learn. 
We balanced our training set by randomly down-sampling the no-stress class depending on the number of stress samples annotated for each participant.
In addition to the baseline models we also trained a simple LSTM network on the autoencoder features extracted from the eye tracker video data. The model consists of one LSTM layer with a time step size of 50 frames and one fully connected layer. Unfortunately, some participants accidentally manipulated the eye tracker and changed the alignment of the built-in camera. In some cases, this resulted in uncaptured eyes. 
Therefore the LSTM network could only be trained on a subset of the recorded data. The subset contains 19 sessions for training the model. For that reason, we report the results separately from the baseline results. Apart from the reduced training data, the procedure for training the LSTM model was similar to the other classifiers.

\subsubsection{Evaluation Metrics}
Similar to previous work~\cite{schmidt2018wesad}, we use accuracy and f1-score as the performance metrics to evaluate our stress models.
To assess the generalizability of our models on data from unseen users, we perform LOSO (leave-one-subject-out) evaluations.

\section{Results}

\subsection{Automatic Stress Detection}
We evaluate our dataset on a binary stress recognition task (stress vs. no stress).
Popular machine learning techniques such as RFC, KNN, SVM, and simple feed forward neural networks are trained on features extracted from facial action units, EDA, HRV, OpenPose, and Gemaps.
The results of our LOSO evaluation are presented in~\autoref{tab:classification}.

\begin{table*}[ht]
\caption{Evaluation of classifiers on different modalities for the binary stress recognition task}
\begin{center}
\begin{tabular}{|l|l|l|l|l|}
\hline
 Features & RFC & KNN & SVM & Simple NN \\ \hline
Action Units & F1 = 71.4, Acc. = 73.6 & F1 = 70.7, Acc. = 73.1 & F1 = 75.6, Acc. = 77.2 & F1 = 76.5, Acc. = 78.0 \\
EDA & F1 = 54.2, Acc. = 57.1 & F1 = 54.6, Acc. = 55.2 & F1 = 57.6, Acc. = 58.9 & F1 = 60.2, Acc. = 61.3 \\
HRV & F1 = 74.5, Acc. = 75.9 & F1 = 72.5, Acc. = 73.2 & F1 = 76.1, Acc. = 77.7 & F1 = 78.4, Acc. = 79.7 \\
OpenPose & F1 = 59.4, Acc. = 63.6 &  F1 = 67.0, Acc. = 69.5 & F1 = 69.8, Acc. = 73.4 & F1 = 76.4, Acc. = 79.5 \\
Gemaps &  F1 = 52.1, Acc. = 55.9 & F1 = 55.1, Acc. = 56.9 & F1 = 57.3, Acc. = 58.9 & F1 = 58.7, Acc. = 60.3 \\
\hline
All modalities (early PCA) &  F1 = 81.3 , Acc. = 82.0 & F1 = 74.7, Acc. =  75.5 & F1 = 83.8, Acc. =  84.5 & F1 = 88.1 , Acc. = 88.3  \\
All modalities (late PCA) &  F1 = 78.2 , Acc. = 79.3 & F1 = 74.4 , Acc. = 75.2 & F1 = 83.9, Acc. = 84.5 & F1 = 87.5, Acc. = 87.7  \\
\hline
\end{tabular}
\label{tab:classification}
\end{center}
\end{table*}

Combining modalities (both early and late PCA) yields better stress recognition performance than individual modalities.
However, early PCA achieves slightly better performance across classifiers.
The best stress recognition performance ($F1 = 88.1\%$, $Accuracy = 88.3\%$ ) is obtained by a simple feed-forward neural network using all modalities with early PCA.

The simple feed-forward neural networks consistently outperform other models across modalities.
This is in line with the observations of related work~\cite{bobade2020stress} that used simple neural networks on other stress datasets.

When considering stress recognition using a single modality, HRV features yield the best results across classifiers, followed by facial action units and OpenPose features.
The Gemaps (speech) and EDA features rank the lowest in stress recognition performance, achieving $15 - 20\%$ lower f1-score and accuracy.

As mentioned in \autoref{automatic_stress_detection} we also trained a simple LSTM network on the extracted eye autoencoder features. The model achieved a f1-score of 68.3\% and an accuracy of 70.2\%. 

\subsection{Biological Stress}
As a manipulation check, i.e. to prove whether our job interview scenario indeed induced stress, biological and perceived stress were measured at 6 time points (2 before and 4 after the job interview). 
Cortisol levels as a marker for biological stress significantly changed during the whole session (\autoref{fig:saliva_samples_timecourse}A; \textit{F}(5, 190) = 3.19, \textit{p} = 0.009). They were highest 5 minutes after the job interview and then decreased to baseline levels 35 minutes after the stressor. 
A similar time course was found for perceived stress, which was highest immediately after the job interview and decreased to baseline afterwards (\autoref{fig:saliva_samples_timecourse}B; \textit{F}(5, 190) = 39.82, \textit{p} $< $ 0.001) 

\begin{figure}[ht]
\centering
\includegraphics[width=0.49\textwidth]{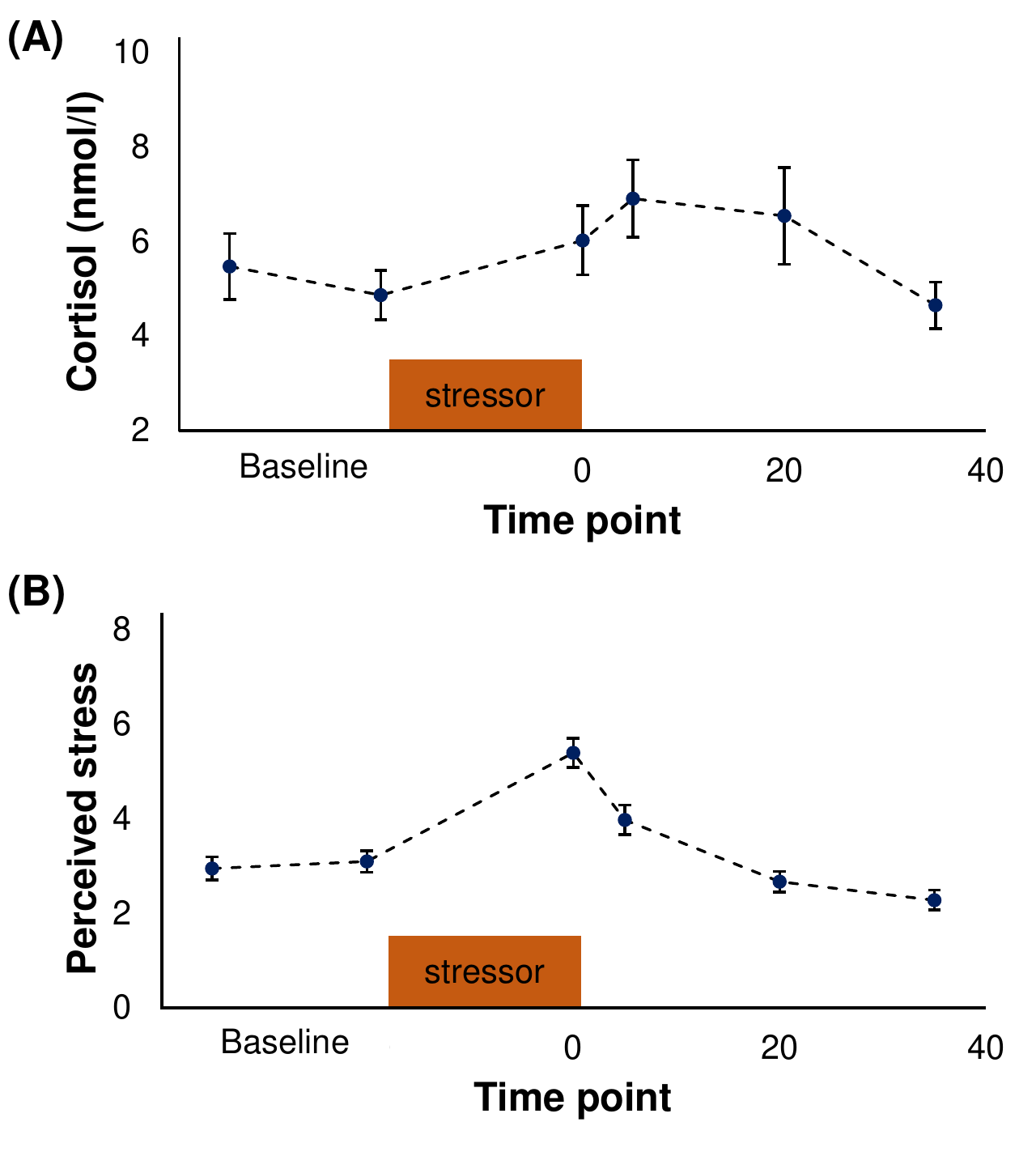}
\caption{Time course of cortisol levels (A) and perceived stress (B) during the whole session.}
\label{fig:saliva_samples_timecourse}
\end{figure}

\section{Discussion}
In order to establish a baseline on our dataset for the automatic recognition of stress we trained several machine learning models on different modalities. Throughout our experiments, a simple NN performed best across all modalities. In single-modality stress recognition, the models trained on HRV features achieved the best results. This is in line with existing research that identified heart rate and HRV as excellent measures for predicting stress \cite{automatic_stress_review, kim2018stress_hrv}. Moreover, models trained with action units and OpenPose features achieved similar results, i.e., 78.0\% and 79.5\% compared to 79.7\% for the HRV features. Another well-established modality to detect stress is EDA \cite{automatic_stress_review}. Models solely trained on EDA features were able to achieve accuracy scores of up to 91\% in a binary stress recognition task \cite{eda_stress_detection}. Interestingly, in our experiments, the models trained on the EDA features were the ones having the second worst accuracy and f1-scores. One reason for that observation could be that existing datasets often aggregate larger time frames to one label whereas we worked with time-continuous annotations with a high temporal resolution. This could be a problem when working with EDA as there is a delay from the sympathetic nervous systems stimulation and the corresponding EDA response \cite{eda_delay}. Therefore, the EDA features could still represent a non-stressed state due to the delay for situations identified as stress. This could potentially be mitigated by either shifting the signal corresponding to the delay or calculating the EDA features over a longer time window. Further investigations should be conducted in future work to check whether following those approaches leads to better classification performance.
The worst-performing modality in our experiments was the GEMAPS features. The best classifier only achieved an accuracy of 60.3\%. Similar research \cite{han2018stressspeechdeeplearning}\cite{kurniawan2013stressspeech} reported slightly better classification accuracies of 66.4\% and respectively 70.0\%. Especially subject independent stress classification based on audio features shows room for improvement when compared to other modalities. Investigating other deep learning architectures like LSTM models or CNN models trained on spectrograms to automatically detect stress could be promising.
The model solely trained on the eye features achieved in our experiment an accuracy of 70.2\%. This places the model in the midfield regarding single modality classification performance even though the model could only be trained on a subset of the recorded data. However, this was the only model that has been trained on time series data therefore results can not be compared without further ado. Nevertheless, the results indicate that features extracted from close-up eye video data hold relevant information for the recognition of stress. Considering that there is only very limited research available \cite{heimerl2022pupillometry} that has been using close-up eye features to automatically detect stress this experiment highlights the usefulness of such features. Features derived from the movement of the eye as well as changes in pupil size are a promising, non-invasive modality for the automatic recognition of stress.
Overall we found that a fusion of the action unit, EDA, HRV, OpenPose and GEMAPS features that already have been reduced in dimensionality by employing PCA achieved the best accuracy and f1-scores with 88.3\% and 88.1\%. 

In order to validate whether mock digital job interviews are a suitable scenario for inducing stress we measured biological as well as perceived stress during the study. Salivary cortisol levels were used as a marker for biological stress. We found a significant change in cortisol levels and perceived stress throughout the study. Peak cortisol levels were observed 5 minutes after the interview whereas perceived stress was found to be highest immediately after the interview. The delay of peak cortisol levels in comparison to perceived stress ratings is due to the fact that it takes some time for the body to release cortisol. In order to reach peak cortisol levels it usually takes 10 to 30 minutes \cite{bozovic2013salivarycortisol}. This delay can be observed in \autoref{fig:saliva_samples_timecourse}. Overall, the results show that mock digital job interviews are a reliable scenario to induce stress in participants. Finally, the stress response can be associated with a variety of person characteristics such as personality or coping styles. Whether the biological stress response in our study was associated with person variables has been analyzed and is reported in \cite{becker2023_fordigitstress}.

\section{Conclusion}
In this paper, we present a comprehensive multi-modal stress dataset that is employing a digital job interview scenario for stress induction. The dataset provides signals from various sources including audio, video, body skeleton, facial landmarks, action units, eye tracking, physiological information (PPG, EDA), as well as already extracted features like GEMAPS, OpenPose, pupil dilation and HRV. In total, 40 participants have been recorded resulting in approximately 56 hours of multi-modal data. Moreover, the dataset contains discrete annotations created by two experienced psychologists for stress and emotions that occurred during the interviews. The inter-rater reliability for the individual stress and emotion labels showed a substantial to almost perfect agreement (Cohen's $\kappa > 0.7$ for all labels). Based on the stress annotations several machine learning models (SVM, KNN, NN, RFC) were trained to predict stress vs. no-stress. The best single modality performance of 79.7\% was achieved by a NN trained on the HRV features. The best stress recognition performance ($F1 = 88.1\%$, $Accuracy = 88.3\%$ ) was obtained by training a NN on all modalities with early PCA. 

Moreover, we validated whether the digital mock job interviews are capable of inducing stress by assessing salivary cortisol levels and perceived stress. The analysis revealed a significant change in cortisol levels and perceived stress throughout the study. Therefore, we conclude that digital mock job interviews are well-suited to induce biological and perceived stress.

In summary, the dataset presented in this work provides the research community with a comprehensive basis for further experiments, studies, and analyses on human stress.

In future work, we plan to establish an additional baseline for the automatic detection of emotions that occurred during the interviews. For this purpose, we plan to extend the dataset by valence and arousal annotations.

\section{Ethics}
The study has been approved by the local Ethics Committee of the FAU (protocol no.: 21-408-S). All participants gave written and informed consent for participation and for publication of their data. Moreover, the presented study has been approved by the data protection officer of the University of Augsburg.

\section*{Acknowledgements}
This work presents and discusses results in the context of the
research project ForDigitHealth. The project is part of the Bavarian Research Association on Healthy Use of Digital Technologies and Media (ForDigitHealth), which is funded by the Bavarian Ministry of Science and Arts. Linda Becker was funded by the Emerging Talents Initiative of the Friedrich-Alexander-Universität Erlangen-Nürnberg. 
We thank Leonie Bast, Steffen Franke, and Katharina Hahn for data collection. 

\bibliographystyle{IEEEtran}
\bibliography{main}

\end{document}